\documentclass{article}

\PassOptionsToPackage{numbers, compress}{natbib}

\usepackage[final]{nips_2018}

\usepackage[utf8]{inputenc} 
\usepackage[T1]{fontenc}    
\usepackage{hyperref}       
\usepackage{url}            
\usepackage{booktabs}       
\usepackage{amsfonts}       
\usepackage{nicefrac}       
\usepackage{microtype}      

\usepackage[svgnames]{xcolor} 
\usepackage{multirow}
\usepackage{amsfonts, amsmath, amsthm, amssymb} 
\usepackage{bbm}
\usepackage{graphicx} 

\title{Phenotype Inference with Semi-Supervised Mixed Membership Models}

\author{
  Victor A. Rodriguez\\
  Department of Biomedical Informatics\\
  Columbia University\\
  New York, NY 10025\\
  \texttt{victor.a.rodriguez@columbia.edu}\\
  \And
  Adler Perotte\\
  Department of Biomedical Informatics\\
  Columbia University\\
  New York, NY 10025\\
  \texttt{adler.perotte@columbia.edu}\\
}

\begin{document}

\maketitle

\begin{abstract}
  Disease phenotyping algorithms process observational clinical data to identify patients with specific diseases. Supervised phenotyping methods require significant quantities of expert-labeled data, while unsupervised methods may learn non-disease phenotypes. To address these limitations, we propose the Semi-Supervised Mixed Membership Model (SS3M) -- a probabilistic graphical model for learning disease phenotypes from clinical data with relatively few labels. We show SS3M can learn interpretable, disease-specific phenotypes which capture the clinical characteristics of the diseases specified by the labels provided.
\end{abstract}

\section{Introduction}

  Phenotypes are powerful tools for working with observational clinical data in the absence of reliable disease labels \citep{hripcsak-phen}. Disease-specific phentoypes allow researchers to sift through large-scale clinical data stores to identify patients with evidence of specific clinical conditions. By answering the question of who has what disease, phenotypes power essential tasks such as cohort selection, trial recruitment and clinical outcome prediction \citep{hripcsak-phen,richesson-phen1,richesson-phen2,pathak-phen}.

  Traditionally, phenotypes were developed by groups of clinical experts who painstakingly hand-tuned rule-based algorithms. The limited scalability of this approach has led to the development of automated methods for learning phenotypes directly from clinical data. Many studies in this vein utilize supervised machine learning methods to build phenotyping algorithms \citep{bergquist-lc,esteban-dm}. Though this approach avoids laborious expert knowledge engineering, it requires significant amounts of labeled clinical data generated by manual chart review. 

  To avoid costly, expert-generated disease labels, many authors have utilized unsupervised methods to cluster patients according to underlying patterns in their clincal data \citep{joshi-nnmf,ho-nntf1,ho-nntf2,wang-rubik,miotto-dp}. In this setting, such patterns play the role of phenotypes. Unsupervised phenotyping methods often learn multiple phenotypes simultaneouly, which may confer evidence of specific diseases. However, such phenotypes generally are not guaranteed to represent single disease concepts. This complicates their evaluation and use in downstream tasks.

  In this paper we propose the Semi-Supervised Mixed Membership Model (SS3M), a probabilistic graphical model which utilizes relatively few disease labels to learn multiple disease-specific phenotypes from observational clinical data. SS3M addresses the limitations of supervised phenotyping by reducing the amount of labeled data needed to learn disease phenotypes; patients do not need to have labels for all diseases. SS3M also addresses the limitations of unsupervised phenotyping by associating disease labels with the phenotypes to be learned; a label specifies which disease a phenotype is meant to represent. This simplifies the evaluation of SS3M phenotypes.

  Our overall goal is to utilize SS3M in an active learning framework, where we may request small batches of labels from clinical experts as needed to learn a phenotype for a given disease. In the present work, we aim to evaluate SS3M's capacity for learning disease-specific phenotypes by utilizing readily available ICD9 diagnosis codes as our labels. We consider this phase in our work a crucial step towards obtaining a reliable semi-supervised model.

\section{Semi-Supervised Mixed Memberships Models}

  SS3M is closely related to the popular topic model Latent Dirichlet Allocation (LDA) \citep{blei-lda}, and its multi-modal \citep{pivovarov-uphenome} and supervised \citep{ramgage-llda} extensions. We detail the generative process for SS3M and provide a graphical model in the appendix.

  \paragraph{Inference} We implement a Gibbs sampler to sample latent variables from SS3M's posterior distribution. Conjugacy gives us the complete conditional distributions for the patient-phenotype distributions, $\theta_d$, and phenotype-token distributions, $\phi_{sp}$, in closed form. The complete conditionals for phenotype assignments, $z_{sdn}$, and phenotype activations, $A_{dp}$, are easily normalized. We use Hamiltonian Monte Carlo to sample from the complete conditionals of $B_p$ and $B^*$ \citep{neal-hmc}. Setting the path length $L=25$ and step size $\epsilon=0.01$ yielded stable trajectories with high acceptance rates in preliminary experiments.

\section{Experiments}

  \subsection{Dataset}

    We train all our models using clinical data extracted from the Medical Information Mart for Intensive Care version III (MIMIC-III) \citep{johnson-mimic}. Our dataset is restricted to the first hospital admission of 46,520 neonatal and adult patients. Clinical notes, labs, and medications comprise corpora of clinical observations. The 50 most common ICD9 diagnosis codes form our label set. Tokenized clinical observation corpora were preprocessed to remove stopwords as well as low- and high-frequency tokens. The final preprocessed corpora and labels were then split into training and test sets containing data for 80\% and 20\% of patients respectively.

  \subsection{SS3M}

    We train SS3M to learn $P=70$ phenotypes; 50 labeled phenotypes (one for each ICD9 diagnosis code label) and 20 unlabeled phenotypes. In preliminary experiments, we found the addition of unlabeled phenotypes improved the overall interpretability of their labeled counterparts.\newline
    For a given patient, the model may treat the absence of a label in two ways. In the first, a missing label indicates that the label's activation, $A_{dp}$, should be set to 0. In the second, an absent label indicates the corresponding activation should be estimated. This latter case is meant to model the uncertainty associated with assignment of diagnosis codes in clinical settings. We train seperate models to explore both cases.\newline
    During inference, we have the option to sample the $B_p$ and $B^*$ or hold them fixed. Preliminary experiments suggest holding these variables fixed results in more interpretable labeled phenotypes. We explore both cases initializing the $B_p$ and $B^*$ with draws from Gamma distributions with shape and scale parameters $(10.0, 1.0)$ and $(0.01, 1.0)$ respectively.\newline
    In all experiments we set $\alpha = 0.1$ and $\gamma_s = (.01, ..., .01)$, and sampled latent variables for 200 iterations.

  \subsection{Quantitative evaluation} 

    In all cases involving mixed membership models, global latent variables are learned on the training set. Trained global variables are then loaded into models exposed to patient data in the test set.

    \paragraph{Label prediction}
    For SS3M, we estimate each test patient's posterior likelihood for each disease label. We evaluate against logistic regression (LR) and naive Bayes (NB) models trained with raw tokens or patient-phenotype distributions learned with either of two mixed membership models. The first of these is the multi-channel mixed membership model (MC3M) -- an unsupervised model similar in structure to SS3M but with a simple Dirichlet prior on the patient-phenotype distributions, $\theta_d$ \citep{pivovarov-uphenome}. The second is SS3M trained without labels. This model is effectively MC3M with a structured prior (MC3M-SP) on the $\theta_d$.

    \paragraph{Log-likelihood}
    We report the maximum complete data log-likelihoods observed during training of each of our three mixed membership models: SS3M, MC3M, and MC3M-SP.

  \subsection{Qualitative evaluation}

    Here we leverage the judgement of a clinical expert to asses the quality of SS3M phenotypes relative to MC3M phenotypes. In particluar, we aim to evaluate the \textit{coherence}, \textit{granularity}, and \textit{label quality} of learned phenotypes. We asked our clinical expert to complete the following tasks, which were inspired by the expert evaluations detailed in Pivovarov et al. \citep{pivovarov-uphenome}

    \paragraph{Coherence}
      A coherent phenotype was defined as a phenotype containing observations typical of a single disease while omitting observations atypical of said disease. The clinical expert was asked to rate the coherence of individual phenotypes using a five-point Likert scale, with 1 and 5 signifying low and and high coherence respectively.

    \paragraph{Granularity}
      We defined three categories of phenotype granularity: \textit{(1)} non-disease, \textit{(2)} mixture of diseases, \textit{(3)} single disease. We asked our expert to assign each phenotype to one of these categories.

    \paragraph{Label quality}
      We asked our clinical expert to generate a label for each phenotype. If no such label came to mind, the expert was asked to omit this step. If the phenotype in question was learned using SS3M, the expert was asked if their label was equivalent to the phenotype's true label. In addition, the expert was asked to specify how well the true label matched its learned phenotype using a five-point Likert scale with 1 indicating no match and 5 a perfect match.

    The phenotypes for our qualitative evaluations were learned using SS3M and MC3M models with $P=70$. The 50 labeled SS3M phenotypes and 50 randomly chosen MC3M phenotypes were shuffled together. During the evaluation, a single phenotype was drawn and shown to our expert who then completed the tasks above. This was repeated until all 100 phenotypes had been evaluated.

\section{Results}

\begin{figure}[ht!]
    \centering
    \includegraphics[width=1.0\linewidth]{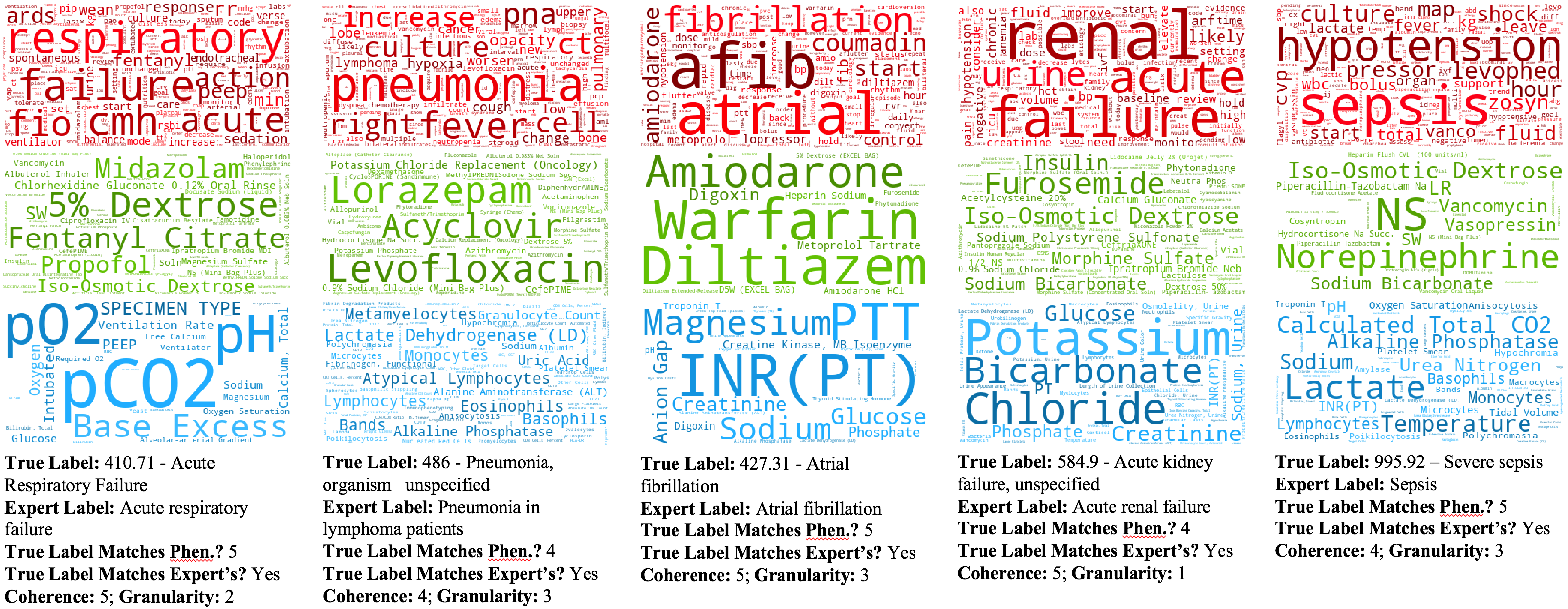}
    \caption{Sample of SS3M phenotypes with their true labels (ICD9 diagnosis codes), expert generated labels, and qualitative evaluation results. \color{Red} Red tokens \color{Black} (top row): words from clinical notes. \color{Green} Green tokens \color{Black} (middle row): clinical labs. \color{Blue} Blue tokens \color{Black} (bottom row): medications. A token's size represents its probability relative to other tokens under the phenotype.}
    \label{sample_ss3m_phen}
  \end{figure}

  Our clinical expert judged SS3M phenotypes to have higher coherence and granularity relative to MC3M phenotypes. Though less than a quarter of the expert's labels were found to match the true labels for SS3M phenotypes, nearly half (48\%) of these phenotypes were found to be good matches for their true labels (Likert score $\geq 3$). A summary of the results from our qualitative evaluation is presented in Table \ref{qual_eval_sum}. A sample of phenotypes judged to match both expert and true labels are shown in Figure \ref{sample_ss3m_phen}.

  Nearly all multilabel classification baselines displayed superior performance relative to SS3M in micro and macro averaged AUROC and AUPRC. Overall, logistic regression trained on raw tokens displayed the best performance. Notably, among SS3M models, the simplest model holding missing label activations fixed (fix $A_0$) and fixing $B_p$ and $B^*$ (fix $B/B^*$) was best in micro and macro averaged AUROC. Meanwhile, the SS3M model with best micro and macro averaged AUPRC had fixed missing activations and sampled $B_p$ and $B^*$ (smpl $B/B^*$).

  The overall best log-likelihood was obtained with our most flexible mixed membership model: SS3M with sampled activations and sampled $B_p$ and $B^*$. A summary of the results from our quantitative evaluation is presented in Table \ref{quant_eval_sum}

  \begin{table}[ht!]
    \caption{Qualitative Evaluation Summary}
    \label{qual_eval_sum}
    \resizebox{\textwidth}{!}{%
      \begin{tabular}{c c c c c c c c c c c c c c c c}
      \toprule
      {} & \multicolumn{5}{c}{Coherence} & \multicolumn{3}{c}{Granularity} & \multicolumn{5}{c}{True label matches phen.?} & \multicolumn{2}{c}{True label matches expert's?}\\
      \cmidrule(lr){2-6} \cmidrule(lr){7-9} \cmidrule(lr){10-14} \cmidrule(lr){15-16}
      {}   & 1  & 2  & 3  & 4  & 5  & 1  & 2  & 3  & 1  & 2  & 3  & 4  & 5  & Y & N\\
      \midrule
      SS3M & 20\% & 8\%  & 14\% & 36\% & 22\% & 24\% & 32\% & 44\% & 36\% & 16\% & 12\% & 22\% & 14\% & 26\% & 74\%\\
      MC3M & 18\% & 28\% & 32\% & 20\% & 2\% & 42\% & 42\% & 16\%  & -- & -- & -- & -- & -- & --   & --\\
      \bottomrule
      \end{tabular}
    }
  \end{table}

  \begin{table}[ht!]
    \caption{Quantitative Evaluation Summary}
    \label{quant_eval_sum}
    \resizebox{\textwidth}{!}{%
      \begin{tabular}{c c c c c c c c c c c c}
        \toprule
        {} & {} & \multicolumn{4}{c}{SS3M} & \multicolumn{2}{c}{MC3M-SP} & \multicolumn{2}{c}{MC3M} & \multicolumn{2}{c}{Raw Tokens}\\
        \cmidrule(lr){3-6} \cmidrule(lr){7-8} \cmidrule(lr){9-10} \cmidrule(lr){11-12}
        {} & {} & smpl $A_0$   &  smpl $A_0$ & fix $A_0$   & fix $A_0$   & \multirow{2}{*}{LR}    & \multirow{2}{*}{NB}   & \multirow{2}{*}{LR}    & \multirow{2}{*}{NB}   & \multirow{2}{*}{LR}    & \multirow{2}{*}{NB} \\
        {} & {} & smpl $B/B^*$ &  fix $B/B^*$ & smpl $B/B^*$ & fix $B/B^*$  & {} & {} & {} & {} & {} & \\

        \cmidrule(lr){3-3} \cmidrule(lr){4-4} \cmidrule(lr){5-5} \cmidrule(lr){6-6} \cmidrule(lr){7-7} \cmidrule(lr){8-8} \cmidrule(lr){9-9} \cmidrule(lr){10-10} \cmidrule(lr){11-11} \cmidrule(lr){12-12}
        \multirow{2}{*}{AUROC} & micro & 0.639   &  0.653 & 0.719             & \underline{0.720} & 0.865 & 0.805 & 0.837 & 0.826 & \textbf{0.884} & 0.803\\
        {}                     & macro & 0.595   &  0.647 & 0.695             & \underline{0.714} & 0.810 & 0.763 & 0.779 & 0.778 & \textbf{0.840} & 0.780\\
        \midrule
        \multirow{2}{*}{AUPRC} & micro & 0.187   &  0.156 & \underline{0.299} & 0.249             & 0.427 & 0.283 & 0.353 & 0.308 & \textbf{0.527} & 0.309\\
        {}                     & macro & 0.162   &  0.187 & \underline{0.252} & 0.245             & 0.277 & 0.215 & 0.232 & 0.229 & \textbf{0.426} & 0.236\\
        \midrule
        \midrule
        \multicolumn{2}{c}{Log-likelihood} & $\mathbf{-7.075 \times 10^8}$ &  $-7.395 \times 10^8$ & $-8.457 \times 10^8$ & $-8.420 \times 10^8$ & \multicolumn{2}{c}{$-7.389 \times 10^8$} & \multicolumn{2}{c}{$-9.623\times10^8$} & \multicolumn{2}{c}{--}\\
        \bottomrule
      \end{tabular}
    }
  \end{table}

\section{Discussion}
  
  SS3M is a model constructed for learning disease phenotypes from observational clinical data in a semi-supervised or active learning setting. As a preliminary step toward this goal, we have exposed the model to MIMIC-III clinical data and derived phenotype labels from common ICD9 diagnosis codes. In quantitative evaluations, SS3M was generally outperformed by baselines in predicting disease labels on held out patient data. Meanwhile, in qualitative evaluations conducted by a clinical expert, SS3M phenotypes were judged to be more coherent, and more granular than phenotypes learned with a related unsupervised model. Moreover, SS3M phenotypes were often found to capture the clinical characteristics of the clinical entities specified by their associated labels. These results suggest that despite its weak performance on label prediction, SS3M does indeed appear capable of learning interpretable, disease-specific phenotypes from clinical data.\newline
  SS3M's weak predictive performance warrants some scrutiny. Suboptimal setting of the model's fixed parameters may be to blame. Indeed, in this work we did little to tune the model's hyperparameters or the total number of phenotypes to be learned. Exploring various settings of these fixed parameters is likely to produce a model with improved predictive performance on held-out test data. It is also possible that our choice of labels is hindering SS3M's performance. ICD9 codes are notoriously noisy proxies for ground-truth diagnosis. This leaves open the possibility that SS3M may actually be generating correct labels for held-out patients who simply lack the appropriate labels for their true underlying clinical conditions. We may better explore this possibility by incorporating SS3M into an active learning framework where high-fidelity labels may be requested as needed from a clinician.
\section{Conclusion}

  SS3M learns disease-specific phenotypes from labeled clinical data. Future work will explore alternative parameterizations to optimize the model's performance on multi-label prediction and utilize higher fidelity label sets. The positive qualitative assessment of SS3M phenotypes encourages us to continue developing SS3M for use in truly semi-supervised or active learning settings.

\section*{Appendix}
  
\begin{figure}[ht]
    \centering
    \includegraphics[width=.9\linewidth]{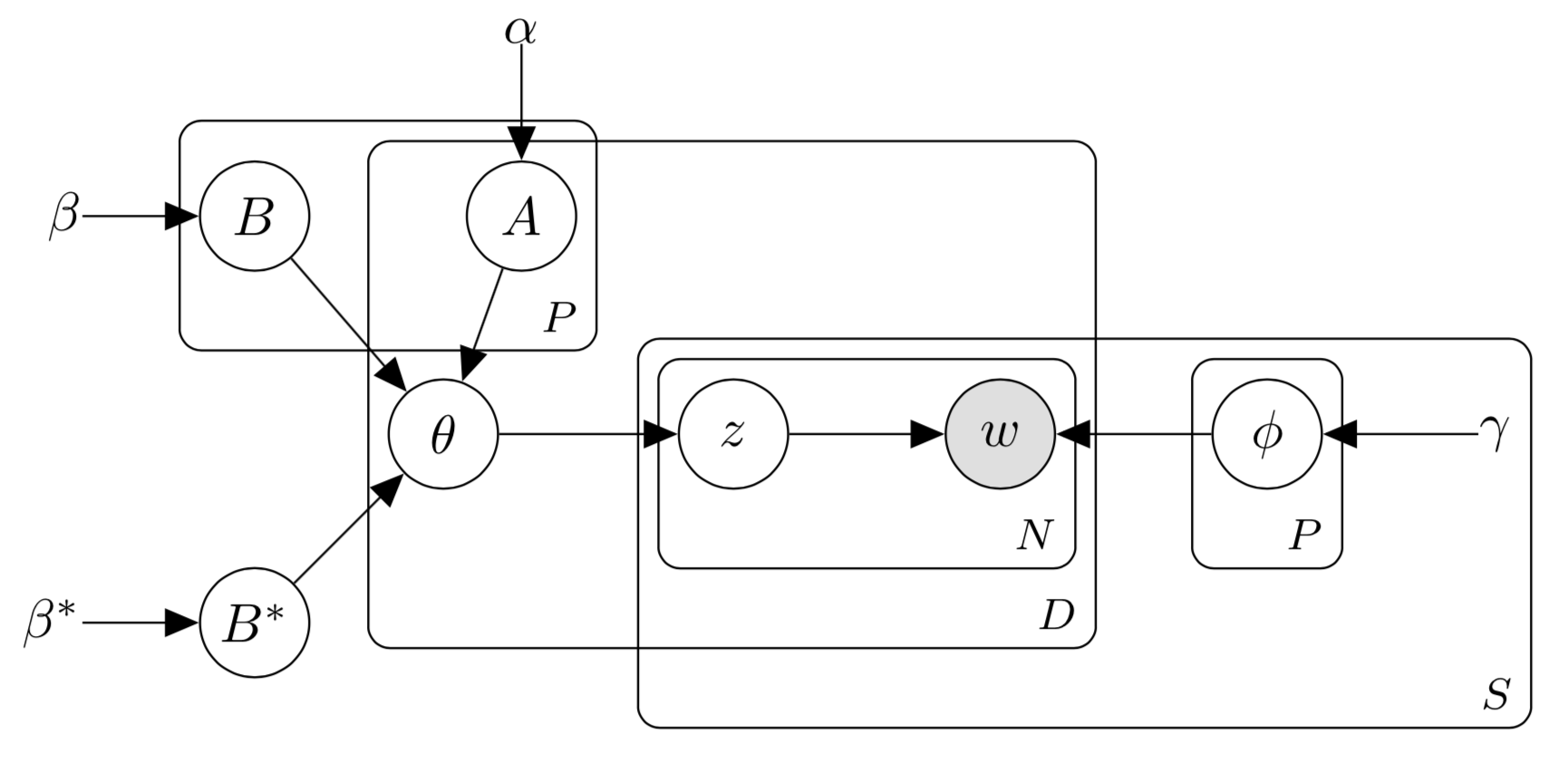}
    \caption{Graphical model for SS3M}
    \label{gm}
  \end{figure}

  The generative process for SS3M is detailed below. The corresponding graphical models appears in Figure \ref{gm}
  
  \begin{enumerate}
    \item For each phenotype $p \in \{1, ..., P\}$ and data source $s \in \{1, ..., S\}$ draw a phenotype-token distribution $\phi_{sp} \sim \text{Dir}(\gamma_s)$.
    \item For each phenotype $p \in \{1, ..., P\}$ draw $B_p \sim \text{Gamma}(\beta_0, \beta_1)$.
    \item Draw $B^* \sim \text{Gamma}(\beta_0^*, \beta_1^*)$.
    \item For each patient $d \in \{1,...,D\}$:
      \begin{enumerate}
        \item For each phenotype $p \in \{1, ..., P\}$ draw phenotype activations $A_{dp} \sim \text{Bern}(\alpha)$
        \item Draw a patient-phenotype distribution $\theta_d \sim \text{Dir}(A_{d,:} \odot B + (\mathbbm{1} - A_{d,:}) B^*)$
        \item For each data source $s \in \{1, ..., S\}$ and observation $n_{sdn} \in \{1,...,N_{sd}\}$ 
        \begin{enumerate}
          \item Draw a phenotype assignment $z_{sdn} \sim \text{Cat}(\theta_d)$
          \item Draw an observed token $w_{sdn} \sim \text{Cat}(\phi_{sz_{sdn}})$
        \end{enumerate}
      \end{enumerate}
  \end{enumerate}

  In steps 2, 3 and 4(a) we introduce patient-level binary phenotype activations $A_{dp}$ and global latent variables $B_p$ and $B^*$ which interact to parameterize the Dirichlet prior on each $\theta_d$ in step 4(b). For patient $d$, if activation $A_{dp}=1$ then the $p^{th}$ parameter to the Dirichlet prior on $\theta_{d}$ is set to $B_{p}$; otherwise, it is set to $B^*$. \newline
  Activations may be estimated, or they may be held fixed in accordance with a set of binary labels reflecting the presence (1) or absence (0) of specific diseases. In this latter case, the presence of a disease label corresponding to phenotype $p$ for patient $d$ would set $A_{dp}=1$ during inference. This constrains the model to use $B_p$ in the prior to $\theta_d'$ instead of $B^*$. The goal of introducing this structure is to encourage the model to learn values for the $B_p$ which push the $\theta_d$ to place most of their mass over phenotypes with positive activations. Thus, by linking activations to labels, a patient's data is driven toward informing inference of phenotypes that capture the distinct clinical meanings associated their labels.

  {}

\end{document}